\def\year{2021}\relax
\documentclass[letterpaper]{article} 
\usepackage{aaai21}  
\usepackage{times}  
\usepackage{helvet} 
\usepackage{courier}  
\usepackage[hyphens]{url}  
\usepackage{graphicx} 
\usepackage{natbib}  
\usepackage{caption} 
\usepackage{times}
\usepackage{adjustbox}
\usepackage{booktabs}
\usepackage{multirow}
\usepackage{latexsym}
\usepackage{boldline} 

\usepackage{xcolor}

\usepackage{amssymb}
\usepackage{amsmath}
\usepackage[utf8]{inputenc}
\usepackage{enumitem}
\usepackage[switch]{lineno}

\DeclareMathOperator{\similarity}{sim}

\title{Extracting Summary Knowledge Graphs from Long Documents}
\author {
        Zeqiu Wu\hspace{10pt}
        Rik Koncel-Kedziorski\hspace{10pt}
        Mari Ostendorf\hspace{10pt}
        Hannaneh Hajishirzi \\
}
\affiliations {
    University of Washington, Seattle, WA, USA \\
    {\tt \{zeqiuwu1, kedzior, ostendor, 	
hannaneh\}@uw.edu}
}

\begin{document}
\maketitle

\begin{abstract}
Knowledge graphs capture entities and relations from long documents and can facilitate reasoning in many downstream applications. Extracting compact knowledge graphs containing only salient entities and relations is important but challenging for understanding and summarizing long documents. 
We introduce a new text-to-graph task of predicting summarized knowledge graphs from long documents.
We develop a dataset of 200k document/graph pairs using automatic and human annotations. 
We also develop strong baselines for this task based on graph learning and text summarization, and provide quantitative and qualitative studies of their effect.

\end{abstract}

\section{Introduction}

Knowledge graphs are popular representations of important entities and their relationships.
Compact, interpretable knowledge can graphs facilitate human data analysis as well as empower memory-dependent knowledge-based applications.
This makes them ideal for modeling the content of documents.
Document-level information extraction which captures relations across distant sentences can be used to construct knowledge graphs of documents \citep{Jia2019DocumentLevelNR, Yao2019DocRED}. 
These techniques focus on extracting all entities and relations from a document, which for long and dense documents such as scientific papers may be hundreds or thousands.
This poses a new challenge: how do we determine the most important entities in a paper and the key relationships between them? 


\begin{figure}[t]
\centering
\includegraphics[width=8.8cm]{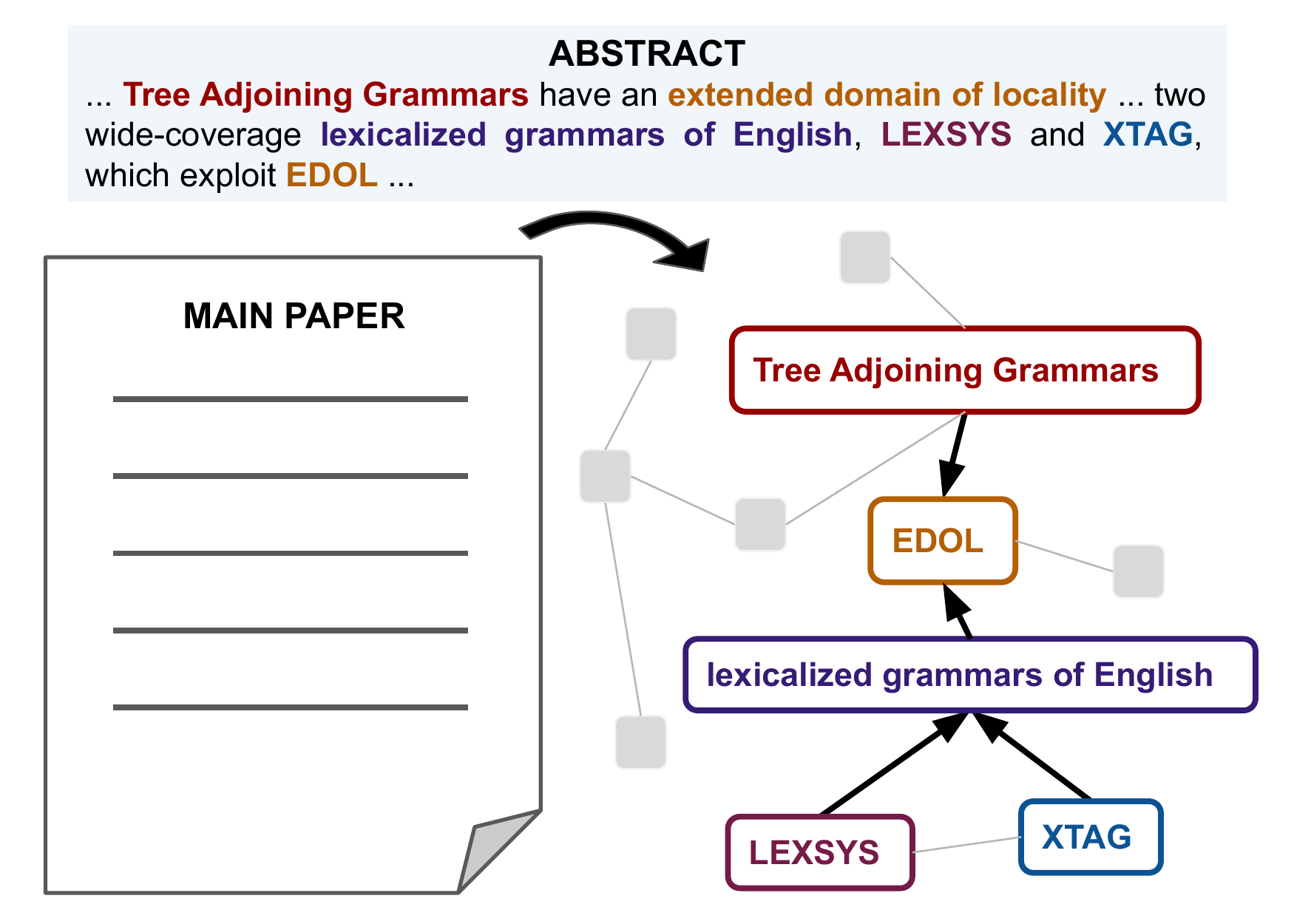}
\caption{\label{fig:task} We introduce the task to extract a summary knowledge graph from a long document (e.g. a scientific paper). This is an example from our dataset, where the target summary graph should only contain entities and relations that are salient enough to be included in the abstract of the paper. The entities or relations (shown in light grey) not found in the abstract should be removed. We omit entity and relation types for simplicity.
}
\end{figure}

Automatic summarization \citep{Liu2019BertSum, Yasunaga2019sci-sum} addresses the problem of identifying salient information in a document, but introduces the additional challenge of discourse structuring (and in the abstractive case, text generation as well).
Summarizing entities and relations directly as a first step could decouple the mixed burdens on models and help assure the factual correctness of a summary in line with recent trends in evaluation for summarization \citep{Wang2020SumQA, Durmus2020FEQA, Zhang2020SumIE}. 


In this work, we introduce the task of extracting from a scientific document a compact knowledge graph that represents its most important information. 
Figure~\ref{fig:task} illustrates the situation: a large knowledge graph can be extracted from the document, but only a portion of entities and relations characterize its main ideas (colored nodes and thick edges), while the rest play a more minor role. 
Our task emphasizes finding this salient subgraph. 
We support this task with a dataset of 200k scientific document/graph pairs that integrates automatic and human annotations from existing knowledge resources in the scientific domain. 
We outline an evaluation paradigm that balances accuracy against redundancy while admitting the variability of textual reference to an entity. 

We develop and investigate two competitive baselines based on text summarization and graph learning models, and compare to two simple frequency-based methods. 
We provide an analysis of their tradeoffs, and of the general challenges posed in the proposed dataset.
For example, we observe that missing entities and entity coreference errors in the predicted graphs have a large impact on relation accuracy.
Our hope is that this task and data will facilitate research into future models that can better capture these challenging but important textual relations.

\section{Background / Related Work}

\subsection{Information Extraction (IE)}

Most IE work focuses on 
extracting entities and their relational facts from a single sentence \citep{Zhang2017tacred, Stanovsky2018OpenIE}. More recent work addresses document-level IE which aims to capture relations across distant sentences \citep{Jain2020SciREXAC, Jia2019DocumentLevelNR, Yao2019DocRED} or to fill a pre-defined metadata table with entities \cite{Hou2019IdentificationOT}. \citet{Jia2019DocumentLevelNR}, \citet{Yao2019DocRED} and \citet{Hou2019IdentificationOT} formulate the task as classifying the relation type between each pair of ground-truth entities expressed in the document. We do not assume the existence of ground-truth entities and extract  document-level relations directly from text.

Our work complements this trend of applying IE for long document understanding while addressing the need for focused, compact knowledge representations. 
The closest work to our is \citet{Jain2020SciREXAC}, who separately explore the idea of identifying salient entities related to experimental results using weak supervision provided by the Papers with Code dataset.\footnote{Papers with Code: \url{paperswithcode.com}} 
In contrast, we focus on identifying salient entities from the paper based on weak supervision from its abstract.
This framing generalizes to a wider variety of documents and domains, and supports diverse tasks including multi-document summarization or building scientific knowledge bases. 
Besides entity salience, our task also requires models to identify the salience of relations, which we show to be challenging.

\subsection{Text Summarization}

Document summarization models create summaries by identifying the most important sentences from documents \citep{Nallapati2017extsum, Narayan2018extsum} or using a decoder to generate abstractive summaries \citep{rush2015abssum, celikyilmaz2017abssum}. Although text summarization tasks \citep{Liu2019BertSum, Yasunaga2019sci-sum} share our objective of distilling crucial information from documents, 
they mix this objective with the goal of producing fluent natural language text. 


We argue that summarizing entities and relations directly as the first step could decouple the mixed burdens on models and help models to check the factual correctness of a summary. These advantages can benefit other text generation tasks that rely on long document understanding and representation, such as generation grounded on long text.
An increasing number of recent works \citep{Wang2020SumQA, Durmus2020FEQA, Zhang2020SumIE} have proposed automatically evaluating summarization models by applying information extraction or question answering models to match entities or relations between generated and reference summaries. These newly proposed measures are found to have much higher correlation with human judgements than standard measures.

Recent applications of large pretrained language models such as \citet{Ribeiro2020InvestigatingPL} and \citet{kale2020texttotext} show the promise of generating fluent and accurate text from knowledge graphs, highlighting the need for identifying correct underlying knowledge representations.
In addition, summarized knowledge graphs from multiple documents can be naturally merged by collapsing shared entity nodes to bring even richer information. And such summarized structures can be more easily leveraged to facilitate downstream tasks.

Another line of work that is closely related to ours is graph-based summarization, which leverages graph structures of documents to facilitate the summary generation \cite{Erkan2004LexRank, Tan2017GraphSum, Yasunaga2019sci-sum, Huang2020GraphSum, Xu2020GraphSum}. These works try to leverage graphs that capture relations between sentences or discourse units. \citet{Wang2020GraphSum} incorporate graphs between entities extracted by sentence-level IE systems without considering entity or relation salience.

\subsection{Graph Summarization}

In general, graph summarization work can be categorized according to whether its goal is optimizing for memory or computational resources needed for processing, or improving analysis \cite{Liu2018Survey}. 
Knowledge graph summarization \citep{safavi2019PerKGSum} or node estimation \citep{park2019KGSum} are most relevant to our work, but they normally take a huge knowledge graph for pruning based on a given query, without any document context.
Unlike these works, our task requires the model to handle different document contexts at inference, where each document can contain a completely different knowledge graph with unique entities.
Our work is also similar to \citet{falke2017concept-map}, who collect a small corpus (30 data instances) of concept map annotations from OpenIE tuples for summarizing sets of documents. We choose a science-specific annotation scheme which provides more structure as our target.

\subsection{Scientific Document Understanding}

Although our proposed idea of summarized knowledge graphs can be applied to documents in any domain, we focus on scientific papers in this work. 
Existing works toward understanding scientific documents include but are not limited to information extraction \citep{Luan2018MultiTaskIO, Wadden2019EntityRA}, summarization \citep{Cohan2018SciSum, Collins2017SciSum, Yasunaga2019sci-sum}, fact verification \cite{Wadden2020fact}, and citation generation \cite{luu2020citation, xing-etal-2020-automatic}. 
Recently, attention to research in the scientific domain in the NLP community has grown even more \citep{Wang2020covid, Esteva2020covid} due to the urgent need for mitigating the COVID-19 global pandemic.

\section{Task Formulation}

\label{section:task}
We introduce a text-to-graph task of extracting a succinct, structured knowledge graph which contains the most salient entities and relations from a document. More specifically, such a summarized knowledge graph should meet these conditions:  1) 
it contains only the most important entities from the document; 2) 
it includes relations between these entities only if they are crucial to understanding the main ideas of the text; and 3)  each salient entity is only represented by a single node in the graph.
These conditions are evaluated as entity salience, relation salience, and entity duplication rate, respectively (evaluation details in Section~\ref{sec:eval}). In our dataset (see Section~\ref{sec:data}) where long documents are scientific papers, the most salient entities and relations are defined to be those that can be included in paper abstracts.


Figure~\ref{fig:task} shows an example of this task. 
Here, only entities and relations that appear in the abstract should be included in the summary knowledge graph. Those in grey should be removed even though they are mentioned in the paper, because they are not necessary enough to describe or understand the main idea of the paper.

Formally, we define the problem as: Given a document $D$, a pre-defined entity type set $T_v$ and a relation type set $T_R$, predict a summarized knowledge graph $\mathcal{G}=(V,E)$, where each $v_i \in V$ represents a salient entity with entity type $t_i \in T_v$ mentioned in $D$. 
Each $(v_i,v_j,r_{ij}^k) \in E$ represents an important edge from $v_i$ to $v_j$ with relation type $r_{ij}^k \in T_R$. 
We note there can be multiple edges between $v_i$ and $v_j$, but $r_{ij}^{k} = r_{ij}^{l} \iff k = l$. 
Each $v_i$ consists of a cluster of $n_i$ string names $\{m_i^1, m_i^2, ..., m_i^{n_i}\}$ (from co-referent entity mentions).

\section{Our Dataset: {\textsc{SciGraphSumm}}} 
\label{sec:data}
We construct a text-to-graph dataset from a corpus of scientific papers. 
Our textual data consists of 
roughly 200k
computer science research papers taken from S2orc, a corpus of 8 million full-text research papers and abstracts \citep{lo-wang-2020-s2orc}. We leverage abstracts to create summarized knowledge graphs for full papers. Abstracts effectively contain summarized information from full documents, and there are existing human annotation and information extraction (IE) systems that enable constructing relational graphs from abstracts.

Due to the expense and difficulty of annotation, we only have access to a small number of human-annotated summary graphs, which we
use to judge system performance. We call this data the ``human test set''. 
For training and development, we take a weakly supervised approach using automatically extracted summary graphs. We use a scientific IE system to extract summary graphs from 196k abstracts and pair them with full papers from S2orc. Entity-relation graphs are also extracted for the full papers. We divide these graph/paper tuples into train, dev and automatic test sets. The dev set can be used for parameter tuning purposes, while the auto test set can be used to compare systems. Randomly sampling 115 examples, we observe that over 90\% of extracted target entities for abstracts in the automatic test set are meaningful. Moreover, we will show in Section~\ref{sec:exp} that similar system performance trends are observed in the ``human test set'' and ``automatic test set''. Table~\ref{data-stat} gives statistics about the number and size of the textual documents, as well as the summary graphs, for all data splits. The data collection and graph construction details are described in the sections that follow. 
The automatic test set is much larger than the human test set, which reduces the problems of noise in automatic annotation.

\begin{table}
\centering
\begin{adjustbox}{width=0.48\textwidth}
\begin{tabular}{c c c c c}
\hlineB{2}
 & Train & Dev & Auto Test & Human Test   \\
\hline
\# examples & 190k & 1k & 5k & 234\\
\# doc tokens & 6.4k & 6.5k & 6.4k & 3.9k \\
\# graph entities & 13.4 & 13.6 & 13.4 & 11.2\\
\# graph relations & 10.9 & 11.1 & 11.0  & 9.1\\
\hlineB{2}
\end{tabular}
\end{adjustbox}
\caption{\label{data-stat} Statistics of each data split.}

\end{table}

\subsection{Manual Summarized Graph Annotation}
We leverage the SciERC scientific IE dataset \citep{Luan2018MultiTaskIO} for the human-labeled test data. 
SciERC consists of 500 expert-annotated paper abstracts, labeled with entities (6 types: Task, Method, Metric, Material, Other Scientific Term and Generic), co-reference, and relations (7 types: Compare, Part-of, Conjunction, Evaluate-for, Feature-of, Used-for, Hyponym-of).
We construct knowledge graphs from these annotations by collapsing coreferent mentions into a single node and linking all nodes via the annotated relations.
Of the 500 abstracts in SciERC, 300 have full texts available in S2orc.
In order to guarantee information richness, We discard pairs where annotated graphs have fewer than 5 predicted relations. After such filtering, our ``human test set'' consists of the knowledge graphs and full text of 234 of these documents.

\subsection{Automatic Summarized Graph Annotation} 
To facilitate model training and model development under a weakly supervised setting, we automatically create target knowledge graphs for the remaining papers from their abstracts.   
We leverage DyGIE++, a state-of-the-art scientific IE system that extracts entities, relations and co-references simultaneously \cite{Wadden2019EntityRA}. We do not re-train DyGIE++, instead we use the pretrained model on SciERC for all processing and modeling steps in this work.
We construct knowledge graphs from the IE output by again collapsing coreferences to create entities, which we associate with the list of coreferential text mentions.
IE relations between mentions become edges between the corresponding entities in the graph. The same sample filtering is applied.\footnote{ We additionally discard pairs with abstracts that are longer than 500 tokens (rare) to avoid memory limitations of DyGIE++.}


\subsection{Automatic Full Graph Construction}
Our task is designed to build a summarized graph directly from a document, and in order to perform the task, models do not have to use specific information extraction tools to build a full graph first. However, we provide the full knowledge graphs that we constructed from documents as part of our dataset for reproducibility and to encourage future exploration on graph learning models for our task.
We process each full document text with DyGIE++ in overlapping 300-token windows (reduce computation memory) with each two consecutive chunks having one overlapped sentence to preserve cross-sentence co-references.\footnote{We discard sentences longer than 150 tokens to guarantee one overlapped sentence between each consecutive chunks (fewer than 1\% sentences discarded).} 
We collapse coreferential mentions as previous steps, and then collapse coreference clusters from different windows with matching unique, non-generic mentions into a single graph node. A non-generic entity mention 
is a string (excluding pronouns and determiners) with more than one token, or a unigram with
an inverse document frequency (IDF) in the training data that is higher than an empirically chosen threshold, 
selected for high precision in identifying generic mentions. A generic entity mention is not clustered unless the model predicts it to be coreferent with some other entity mention.


\section{Evaluation Metrics}
\label{sec:eval}
The goal is to evaluate the correctness of the predicted summary knowledge graph compared with the ground truth summary graph. We first align entity nodes in the predicted graph to nodes in the target graph. After the entity alignment step, we measure 3 qualities of the predicted graphs -- entity salience, relation salience, and duplication rate -- under a relaxed alignment condition, described next.

\subsection{Entity Alignment}
\label{sec:ent_align}
In the ``human test set'', we found 30\% annotated entity mentions do not have exact string match in the main paper text. Further analysis showed that many such cases are due to minor paraphrasing, hyphenation differences or typos caused by OCR parsers that are 1commonly used to process papers in PDF format. For example, \textit{in-domain monolingual corpus} in the abstract and \textit{in domain monolingual corpus} in the paper are equivalent but do not have an exact match due to the hyphen difference. 
In addition, as we do not assume any specific information extraction tools being used, a similar issue for exact name matching may occur, potentially due to different entity mention names being extracted by different models. Therefore, exact string match does not yield a good alignment, and we instead use a relaxed alignment method that we found to be reasonably accurate for evaluation.

Another issue in aligning entities between two graphs is that the same entity can be referred to with multiple strings, as each entity node represents a cluster of co-referent entity mentions. 
%
To align a predicted node with a target node where either can have a cluster of mention types, we find the maximum similarity over all possible pairs. 
The similarity score between a target entity $v_i=\{m_i^1, m_i^2, ..., m_i^{n_i}\}$ and a predicted entity $\hat{v}_j=\{\hat{m}_j^1, \hat{m}_j^2, ..., \hat{m}_j^{{n}_j}\}$ is calculated as:
\begin{equation}
s_{i,j} = \underset{s,t}{\max} \: \similarity(m_i^s, \hat{m}_j^t)
\end{equation}
where we employ ``gestalt pattern matching" \cite{ratcliff1988gestalt} to calculate string similarity based on common substrings.
Each predicted node is aligned with the target node that gives the highest similarity score, subject to a minimum score $\lambda$. $\lambda=0.7$ is selected such that, in a set of 200 relaxed (but not exact) alignments, 90\% of them are manually inspected to be acceptable. 

The 200 manually inspected samples fall into the following categories:
\begin{itemize}
\item \textbf{Paraphrases of Target Nodes (50\%):} We consider relaxed alignment examples to be good if differences only involve typo, hyphen, item order or other paraphrases. For example, \textit{log-linear and linear interpolation} versus \textit{linear and log-linear interpolation}

\item \textbf{Different Specificity Level (40\%):} We consider aligned entities with different specificity level as relevant alignments. For example, \textit{speaker's intention prediction modules} versus \textit{intention prediction modules}.

\item \textbf{Alignments with Error (10\%):} We consider entities being aligned that have distinct meanings to be bad alignments. For example, \textit{two-dimensional analog of sorting} versus \textit{one-dimensional notion of sorting}.
\end{itemize}

In both human and auto test sets, applying the
relaxed alignment from full graph nodes to target nodes increases the percentage of aligned target nodes from 60\% (by exact matching) to 80\%.

\subsection{Salience and Duplication Measures}
To calculate entity salience, we align each predicted node with up to one target graph node, collapsing multiple predicted nodes that map to the same target entity into a single node for calculating precision, recall and F1 scores. In other words, if multiple predicted entities are aligned to the same target entity, it is only counted once when calculating all scores. 
These metrics can be computed either with matching or ignoring entity types  (typed vs.\ untyped evaluation). As each entity should only have one entity type, we adopt the dominating type among all mentions to be the entity type.
Since this process does not penalize predicted graphs where multiple nodes are aligned to a single target node, 
we also calculate the duplication rate as the average number of predicted nodes which are aligned to each target node. 

Based on entity alignment, a target relation edge ($v_i$, $v_j$) can be aligned to a predicted relation for ($\hat{v}_{k}$, $\hat{v}_{l}$) 
if the corresponding nodes align (i.e., $v_i$ aligns to $\hat{v}_{k}$ and $v_j$ aligns to $\hat{v}_{l}$). We evaluate relation salience based on such relation alignments, allowing for multiple relation types between each pair of entities.
We report precision, recall and F1 scores for relation prediction, with or without considering relation type and direction matching  (typed vs.\ untyped evaluation). 
When evaluating without relation type and direction, we merge 
relations (if multiple) between an entity pair into a single edge.

\section{Baseline Models} 
\label{section:model}

We develop two baseline models for the graph summarization problem: one using a text summarization model that extracts summary sentences from which we extract entities and relations, and one that first builds a full-document graph and then applies a graph learning model to do graph pruning.

\subsection{Text-Text-Graph (TTG)} 
This model first produces a text summary of the full document text using the extractive summarizer BertSumExt \cite{Liu2019BertSum}, and subsequently uses entities and relations from the text summary to form a summary knowledge graph. 
We re-train the original model on our dataset, by replacing the pre-trained BERT with SciBERT \cite{Beltagy2019SciBERT} and increase the sequence length from 512 to 1024.
DyGIE++-predicted entities and relations that appear within the text summary are used as the summarized graph.

\subsection{Graph-to-Graph (G2G)}
This model predicts a summary subgraph from the full graph extracted by DyGIE++ (described in Section~\ref{sec:data}).\footnote{We release our data and the model code for G2G at \url{https://github.com/ellenmellon/GraphSum}}

We formulate subgraph selection as a node classification problem: we encode the full graph with a GAT \cite{Velickovic2018GAT} and use the resulting node representations to make a binary salience prediction.

In the original GAT, a node $v_i$ is embedded with a learnable feature vector and contextualized via multi-headed attention with its graph neighbors $\mathcal{N}(v_i)$ in each graph attention layer. At each graph attention layer, a vertex $v_i$ with neighborhood $\mathcal{N}(v_i)$ is contextualized as:
\begin{eqnarray}\label{eq:attn}
    \mathbf{\hat{v}}_i &=& {\bf v}_i + \sum_{j \in \mathcal{N}(v_i)}\alpha_{ij} \mathbf{W}_{V} \mathbf{v}_j \\
     \alpha_{ij} &=& \frac{\exp(({\mathbf{W}_{K}{\bf v}_j)^{\top}\mathbf{W}_{Q}\mathbf{v}_i})}{\underset{z \in \mathcal{N}(v_i)}{\sum}\exp(({\mathbf{W}_{K}{\bf v}_z)^{\top}\mathbf{W}_{Q}\mathbf{v}_i})}
\end{eqnarray}
Here $\mathbf{\hat{v}_i}$, $\mathbf{v_i} \in \mathbb{R}^{h}$ denote contextualized and original vector representations of $v_i$. $\mathbf{W}_V$, $\mathbf{W}_K$, $\mathbf{W}_Q \in \mathbb{R}^{h \times h}$ are model parameters, and $\alpha_{ij}$ are attention weights computed from the vertex representations. 
The formulation above was extended using multi-head attention and layered with non-linearities to produce the Graph Attention Network.

Since the original GAT does not consider different relation types between neighboring vertices, to incorporate relation types into the model, we use separate heads for different relation types in $T_R$; that is, the head corresponding to relation type $R \in T_R$ is used to attend $v_i$ over those $v_j \in \mathcal{N}_R(v_i)$ where $v_i$ and $v_j$ are connected by an edge with label $R$, i.e. $(v_i,v_j,R) \in E$. As we have 7 different relation types in our dataset, we use 7 heads in our GAT.
The representations from all heads are concatenated and transformed via non-linearity between model layers.

At the \textit{node embedding layer}, we use four features to embed each entity node $v_i$: the number of mentions in the document $n_i$, the section id of the entity's first appearance in the document $s_i$, the most frequent entity type among all mentions as predicted by DyGIE++ $t_i$, and the pooled output representation from SciBERT \cite{Beltagy2019SciBERT} of the longest mention string $z_i$, to encode each node as follows:
\begin{equation}\label{eq:node-emb}
\mathbf{v}_i =  n_i\mathbf{n} + \mathbf{W_s}\mathbf{s_i} + \mathbf{W_t}\mathbf{t_i} + \mathbf{W_e}{\it S}(z_i)
\end{equation}

\noindent
where $\mathbf{n} \in \mathbb{R}^{h}$ is a learnable unit feature vector for $n_i$, $\mathbf{s_i} \in \mathbb{R}^{n_s}$ and $\mathbf{t_i} \in \mathbb{R}^{n_t}$ are the one-hot vectors of $s_i$ and $t_i$ respectively, with $n_s$ and $n_t$ as the number of unique section ids and node (entity) types in the dataset. 
$\mathit{S}(z_i) \in \mathbb{R}^{h_e}$ is the hidden representation at the first token ([CLS]) from the final layer of SciBERT.  $\mathbf{W}_s \in \mathbb{R}^{h \times n_s}$ , $\mathbf{W_t} \in \mathbb{R}^{h \times n_t}$, $\mathbf{W}_e \in \mathbb{R}^{h \times h_e}$ are trained model parameters.

Following the node embedding layer, we contextualize each node representation with 6 {\em GAT layers} and pass each node through a final binary classification layer to predict salience.

To supervise the training of this model, we apply the relaxed alignment method (in Section~\ref{sec:eval}) to align full graph entities and target graph entities. 
All full graph entities that can be aligned are treated as positive examples, and all others as negative. Finally, we use a negative log likelihood loss function using all positive (labeled as salient) nodes and a negative sampling ratio of 3 for training.

We include all full graph relations between predicted entities in the output summary graph. We leave better relation prediction for future study.

\paragraph{Implementaion Details} We manually tune the hyperparameters of GAT based on dev set entity F1 performance curve versus training steps. We fix most of the model parameters (e.g. vector dimension $h$ = 16; number of layers = 6; batch size = 10). The only parameters being tuned are learning rate, dropout rate and negative sampling ratio. But we only manually change each parameter value if we observe performance instability on the dev set for the first 1000 training steps. The average number of tuning trials for each parameter is fewer than 5 times. Finally we set negative sampling ratio = 3, dropout rate = 0.2 and learning rate = 5e-5 for all experiments with our G2G model. We use Adam optimizer. We do not finetune SciBERT (the base model) used in GAT. 
We run each experiment on a single TITAN RTX. We select the model checkpoint based on its typed relation F1 score performance on valid set.

\section{Experiments}
\label{sec:exp}

\begin{table*}

\begin{minipage}{1.0\textwidth}
\centering

\medskip

\begin{adjustbox}{width=0.95\textwidth}
\begin{tabular}{c c c c c c c | c c c c c c | c }

\toprule
\multirow{3}{*}{} & \multicolumn{6}{c|}{\bf Untyped} & \multicolumn{6}{c|}{\bf Typed}\\
 & Ent P & Ent R & Ent F1 & Rel P & Rel R & Rel F1 & Ent P & Ent R & Ent F1 & Rel P & Rel R & Rel F1 & E Dup \\
\midrule
PR & 23.6 & 34.3 & 26.8 & 6.6 & 8.1 & 6.7 & 17.2 & 24.8 & 19.5 & 4.8 & 6.5 & 5.0 & 1.30 \\
TKF & 24.6 & 35.4 & 27.9 & 7.8 & 8.1 & 7.3 & 17.8 & 25.5 & 20.1 & 5.5 & 6.4 & 5.3 & 1.34 \\
TTG & {\bf 30.2} & 29.7 & 28.3 & {\bf 11.2} & 6.6 & 7.5 & {\bf 22.7} & 22.3 & 21.2 & {\bf 9.5} & 5.6 & 6.4 & {\bf 1.18}\\
G2G & 29.7 & {\bf 43.8} & {\bf 32.7} & 9.1 & {\bf 12.6} & {\bf 9.2} & 21.6 & {\bf 31.8} & {\bf 23.7} & 6.5 & {\bf 10.2} & {\bf 6.9} & 1.55\\
\midrule
GE & 100.0 & 81.2 & 88.8 & 44.5 & 18.9 & 24.3 & 77.6 & 63.4 & 69.2 & 34.8 & 16.2 & 20.0 & 1.0\\
\bottomrule
\end{tabular}
\end{adjustbox}

\end{minipage}\hfill

\caption{\label{full-results-auto} Full results for \textbf{untyped} and \textbf{typed} entity / relation evaluation (P, R for Precision and Recall) and entity duplication rate (E Dup) on \textbf{auto test} set. All scores are in \%, except entity duplication rate. Note that entity duplication rates are expected to be the same for both untyped and typed evaluation as entity alignment only considers string name matching. }
\end{table*}

\begin{table*}

\begin{minipage}{1.0\textwidth}
\centering

\medskip

\begin{adjustbox}{width=0.95\textwidth}
\begin{tabular}{c c c c c c c | c c c c c c | c }

\toprule
\multirow{3}{*}{} & \multicolumn{6}{c|}{\bf Untyped} & \multicolumn{6}{c|}{\bf Typed} &  \\
 & Ent P & Ent R & Ent F1 & Rel P & Rel R & Rel F1 & Ent P & Ent R & Ent F1 & Rel P & Rel R & Rel F1 & E Dup \\
\midrule
PR & 22.8 & 38.7 & 27.8 & 6.6 & 9.3 & 7.2 & 17.2 & 29.1 & 21.0 & 4.7 & 7.3 & 5.4 &  1.30\\
TKF & 24.3 & 40.7 & 29.4 & 8.9 & 9.4 & 8.4 & 18.0 & 30.0 & 21.8 & 6.1 & 7.1 & 5.9 & 1.41\\
TTG & 33.9 & 35.3 & 32.7 & {\bf 13.7} & 9.3 & 9.9 & 26.1 & 27.0 & 25.1 & {\bf 11.5} & 7.9 & {\bf 8.3} & {\bf 1.17}\\
G2G & {\bf 36.9} & {\bf 42.6} & {\bf 36.9} & 11.3 & {\bf 11.8} & {\bf 10.1} & {\bf 28.0} & {\bf 32.1} & {\bf 27.8} & 8.4 & {\bf 9.3} & 7.6 & 1.42\\
\midrule
GE & 100.0 & 79.4 & 87.6 & 44.5 & 20.3 & 25.8 & 79.9 & 64.0 & 70.3 & 34.8 & 17.3 & 21.3 & 1.0\\
\bottomrule
\end{tabular}
\end{adjustbox}

\end{minipage}\hfill

\caption{\label{full-results-human} Full results for \textbf{untyped} and \textbf{typed} entity / relation evaluation (P, R for Precision and Recall) and entity duplication rate (E Dup) on \textbf{human test} set. All scores are in \%, except entity duplication rate. }
\vspace{-3mm}
\end{table*}


\begin{table}[h]

\centering
\begin{adjustbox}{width=0.45\textwidth}
\begin{tabular}{c c c c c}
\toprule
\multirow{3}{*}{} & \multicolumn{2}{c}{\bf Human test} & \multicolumn{2}{c}{\bf Auto test}\\
 & TTG & G2G &  TTG & G2G   \\
\midrule
Task & 21.9 & 24.4 & 14.3 & 19.1\\
Method & 24.4 & 29.4 &  23.8 & 25.6\\
Metric & 6.5 & 9.7 & 6.3 & 8.9\\
Material & 13.9 & 14.0 & 8.0 & 10.3\\
Other Scientific Term & 17.4 & 20.6 & 17.0 & 20.5\\
Generic & 14.5 & 15.6 & 12.7 & 14.9\\
\bottomrule
\end{tabular}
\end{adjustbox}

\caption{\label{results-entity-type} Entity relaxed match F1 by entity type.}

\end{table}

\subsection{Evaluated Systems} 

We compare the $\textbf{TTG}$ and $\textbf{G2G}$ models in Section~\ref{section:model}
with two frequency-based baselines: $\textbf{PageRank (PR)}$: the top K most ``authorized" entities in the full document graph with the highest PageRank scores \cite{page1999pagerank}, where each edge weight is initialized by the number of relation mentions between the entity pair; $\textbf{TopK-Freq (TKF)}$: the top K most frequent entities. K is selected to be 18 for both of the models, which is the average number of full graph nodes aligned to target nodes in the training set.
In both cases, relations from the full document graphs between the selected entities are added to generate the predicted summary graph.  
We also report performance of $\textbf{Gold Entity (GE)}$, which provides the performance upper bound when relying on the entities and relations that can be extracted from the full text with DyGIE++.
GE picks the full graph node with the highest similarity score for each target node (with a lower threshold of 0.7 for inclusion).
Again, the predicted graph includes all relations found in the full graph between these entities. 

\subsection{Quantitative Results}

Table~\ref{full-results-auto} shows Precision, Recall and F1 scores for both untyped and typed entity and relation prediction, as well as entity duplication rates for the automatic test set. 
We see similar evaluation trends with untyped and typed evaluation, with lower scores when type matching is required, as expected.
%
%
%
%
Both G2G and TTG outperform other baselines in both entity and relation evaluations. In particular, G2G consistently performs the best in F1 scores, though TTG has higher precision. When evaluating entity duplication rate, TTG consistently performs the best. However, all systems' performances are still far from GE, which shows significant room for improvement on this task in the future.

Table~\ref{full-results-human} gives the same results for the human test set.  Most of the trends observed with the automatic test set hold for the human test set. The exception is that G2G shows better performance on entities, but is less effective on typed relations.
%
Therefore, we argue that the automatic test set is reasonable to be used as an extra set to test systems during development.


\subsection{Qualitative Analysis}

We looked at a handful of abstracts in the human test set to analyze the scoring criteria proposed for this task. The target graphs were further hand-annotated to identify the most important entities (among all salient entities in the abstract), eliminating some ``generic'' nodes (e.g.\ ``method''), non-essential ``other scientific term'' nodes, and occasional duplicated nodes. In some cases, we added entities that were not in the gold reference but were identified by one or more of the automatic algorithms and deemed appropriate. Entities identified by the automatic algorithms were hand-aligned to the reduced target set. 
For untyped entities, the recall is higher on the reduced set for G2G and Pagerank, suggesting that these algorithms may be better capturing the most salient entities. Errors in entity types often involved unclear cases. Trends in precision on the reduced graph were consistent with the automatic scoring. What we also noticed is that some of the aligned predicted nodes contain unrelated entities due to coreference errors from IE systems, which in part explains the low relation scores together with the impact of missing/inserted entities.

We investigate the causes of TKF performing poorly compared with G2G and TTG. In specific, we analyze why frequency may not be a good indicator of salient entities in some cases. We first calculate the average length of mention string names of predicted salient entities, and find out TKF has an average length of 2.1 while both TTG and G2G have 2.5 on average. Furthermore, we find out that some important entities tend to have long string names, especially when paper authors start introducing some specific tasks or models. These entities tend to be split into smaller segments in later parts of the paper for more detailed explanations. Such smaller segments tend to be mentioned more frequently and thus predicted by TKF, although sometimes they are not comprehensive enough to qualify as salient entities. For example, a key entity \textit{Bayesian semi-supervised Chinese word segmentation model} can be detected as salient by both G2G and TTG while TKF only predicts \textit{Chinese words} and \textit{word segmentation} as the closest salient entities. Another example is that TKF predicts \textit{KL-One systems} as a salient entity for a paper, while the gold entity is \textit{KL-One-like knowledge representation systems} which is predicted by both G2G and TTG.

As noted in Table~\ref{data-stat}, the sizes of target graphs and full papers are different in the human and automatic test sets.
We observe that TTG produces graphs of similar size for papers in the two test sets: about 12 nodes and 7 edges. By contrast, G2G produces graphs of different sizes for different document lengths, averaging 14 nodes and 10 edges for human test set but 22 nodes and 17 edges for automatic test set, where documents are longer (and target graphs bigger).

We observe interesting trends regarding the sections where entities first appear. 
GE entities appear in the first section of the full paper only 55\% of the time, for both auto and human test sets. About 25\% and 20\% of them have their first mention in middle 5 sections and final sections respectively. These numbers are also consistent with both test sets. This observation highlights the fact that extracting the main idea of the paper needs the understanding of the full paper.
However, partly due to the sequence length limitation of BertSumExt, TTG is extremely biased towards entities in the first section (85\%). 
G2G is less vulnerable to such bias (68\%), but still often fails to include entities from later paper sections in its summaries. 

Table~\ref{results-entity-type} shows the entity F1 score based on relaxed match for each different entity type. We calculate these scores by comparing subgraphs of the predicted and target graphs that contain entities of a certain type only. Entities of type ``Metric" are the hardest to predict. This correlates with the fact that ``Metric'' entities are least likely to appear in the first section of a paper (36.7\% vs 54.9\% overall in human test set). Another possible reason for this is that ``Metric'' is the least frequent salient entity type. Only 5\% and 6\% of all target salient entities have the type ``Metric'', in auto and human test set respectively.


\begin{figure}[h]
\centering
\includegraphics[width=8.5cm]{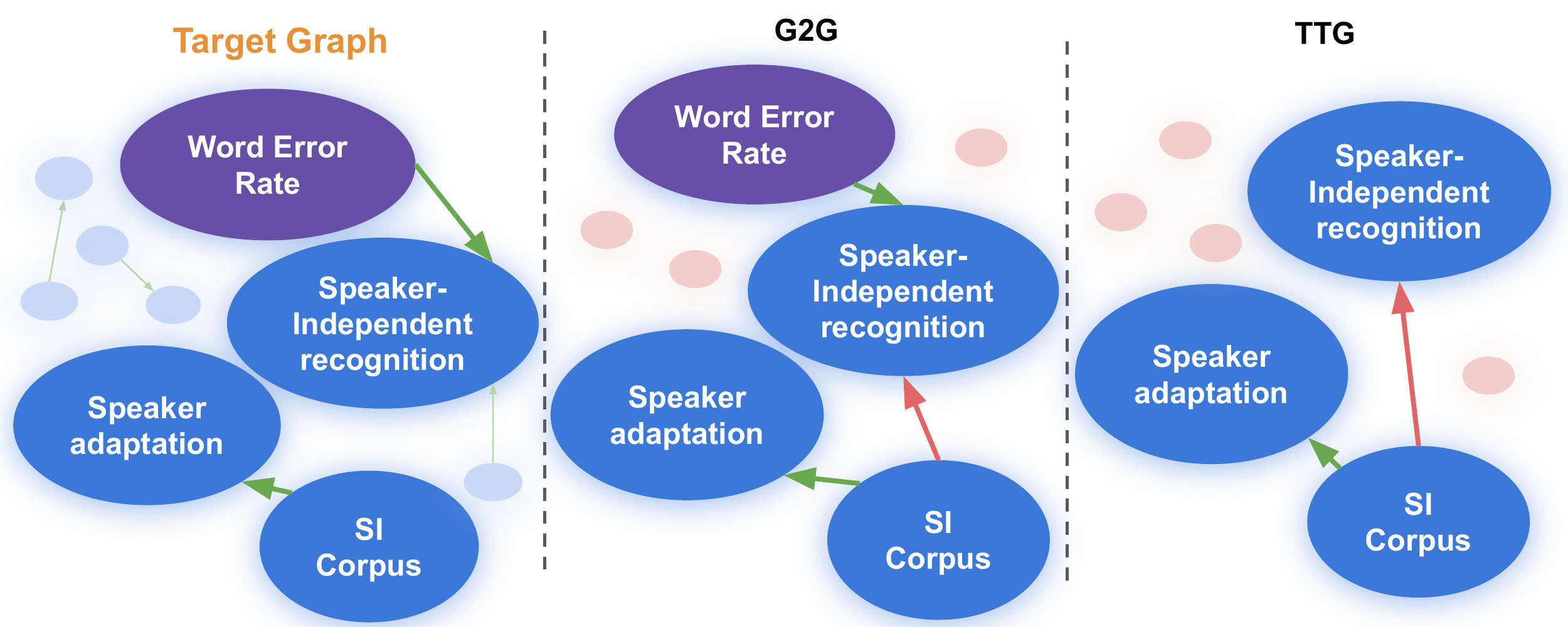}
\caption{\label{fig:case_study_example} Sample output from G2G and TTG. }
\end{figure}
Figure~\ref{fig:case_study_example} shows an example where a target ``Metric" entity first appearing in the second section of a paper is correctly predicted by G2G, but missed by TTG. The red (incorrect) relation edges show that only relying on full graph relations limits relation prediction performance.
This is evidenced in Tables~\ref{full-results-human} and \ref{full-results-auto}, where even GE gives low relation prediction scores due to the absence of gold target relations in the full document graphs. 


\section{Conclusions and Future Work}
We have described a new text-to-graph task for constructing summary knowledge graphs from full text documents, including a standard, preprocessed open-access dataset and evaluation techniques to facilitate further research. 
We have investigated graph classification and text summarization techniques for this task, and detailed some of their qualities in our analysis. 

As we show that relation salience prediction is a rather challenging task in extracting summary knowledge graphs, it would be an important further investigation. Leveraging document-level IE and graph learning techniques would also be an interesting direction to explore. Models that can merge entity nodes better will lead to lower entity duplication rate and improved relation accuracy. One major shortcoming of our GAT model is that we do not consider the context of each entity mention in the document; incorporating contextual information for entity mentions would also be a promising research direction.

\bibliography{aaai21}
\end{document}


\maketitle
\appendix
\section{Appendices}
\label{sec:appendix}

\subsection{Details in Dataset Construction}
\label{sec:explain-dataset}

When constructing knowledge graphs from full text papers, we feed paper sections separately into DyGIE++. Moreover, in order to reduce computation memory, we split long sections into smaller chunks ($\le$ 300 tokens) for processing, with each two consecutive chunks having one overlapped sentence to preserve cross-sentence co-references. We discard sentences longer than 150 tokens (fewer than 1\% cases).
We cluster predicted coreferential entity mentions, as well as entity mentions if they share the same non-generic string name (defined below). Each entity mention with a generic name is not clustered unless the model predicts it to be coreferent with some other entity mention.

\subsection{Relaxed Alignment Examples}
\label{sec:soft-alignment-examples}
The following examples are in the format of aligned ``target graph node string name" VS ``full graph node string name".

\paragraph{Good Alignment Examples}
We consider relaxed alignment examples are good if differences only involve typo, hyphen, item order or paraphrases.
\begin{itemize}[noitemsep]
    \item{} \textit{in-domain monolingual corpus} VS \textit{in domain monolingual corpus}
    \item{} \textit{log-linear and linear interpolation} VS \textit{linear and log-linear interpolation}
    \item{} \textit{unification categorial grammar (ucg)} VS \textit{unification categorial grammar}

\end{itemize}

\paragraph{Relevant Alignment Examples}
We consider aligned entities with different specificity level as relevant alignments.

\begin{itemize}[noitemsep]
    \item{} \textit{shallow techniques} VS \textit{shallow processing techniques}
    \item{} \textit{speaker's intention prediction modules} VS \textit{intention prediction modules}
    \item{} \textit{second language and cognitive skill acquisition} VS \textit{second language acquisition}
\end{itemize}

\paragraph{Bad Alignment Examples}
Bad alignments are entities being aligned that have distinct meanings.

\begin{itemize}[noitemsep]
    \item{} \textit{two-dimensional analog of sorting} VS \textit{one-dimensional notion of sorting}
    \item{} \textit{routing} VS \textit{word counting}
\end{itemize}

\subsection{G2G Model Details}
\label{sec:model-details}

In the original GAT, a node $v$ is embedded with a learnable feature vector and contextualized via multi-headed attention with its graph neighbors $\mathcal{N}(v)$ in each graph attention layer. At each graph attention layer, a vertex $v_i$ with neighborhood $\mathcal{N}(v_i)$ is contextualized as:
\begin{eqnarray}\label{eq:attn}
    \mathbf{\hat{v}}_i &=& {\bf v}_i + \sum_{j \in \mathcal{N}(v_i)}\alpha_{ij} \mathbf{W}_{V} \mathbf{v}_j \\
     \alpha_{ij} &=& \frac{\exp(({\mathbf{W}_{K}{\bf v}_j)^{\top}\mathbf{W}_{Q}\mathbf{v}_i})}{\sum_{z \in \mathcal{N}(v_i)}\exp(({\mathbf{W}_{K}{\bf v}_z)^{\top}\mathbf{W}_{Q}\mathbf{v}_i})}
\end{eqnarray}
Here $\mathbf{\hat{v}_i}$, $\mathbf{v_i} \in \mathbb{R}^{h}$ denote contextualized and original vector representations of $v_i$. $\mathbf{W}_V$, $\mathbf{W}_K$, $\mathbf{W}_Q \in \mathbb{R}^{h \times h}$ are model parameters, and $\alpha_{ij}$ are attention weights computed from the vertex representations. 
The formulation above was extended using multi-head attention and layered with non-linearities to produce the Graph Attention Network.

To incorporate relations into the model, we use a separate head for each relation type; that is, the head corresponding to relation type $R$ is used to attend over those $v' \in \mathcal{N}(v)$ where $v$ and $v'$ are connected by an edge with label $R$, i.e. $(v,v',R) \in E$.
The representations from all heads are concatenated and transformed via non-linearity between model layers. 

At the \textit{node embedding layer}, we use four features to embed each entity node $v_i$: the number of mentions in the document $n_i$, the section id of the entity's first appearance in the document $s_i$, the most frequent entity type among all mentions as predicted by DyGIE++ $t_i$, and the pooled output representation from SciBERT of the longest mention string $z_i$, to encode each node as follows:

\begin{equation}\label{eq:node-emb}
\begin{split}
    \mathbf{v}_i =  & n_i\mathbf{n} + \mathbf{W_s}\mathbf{s_i} \\ 
    &\quad + \mathbf{W_t}\mathbf{t_i} + \mathbf{W_e}{\it SciBERT}(z_i)
\end{split}
\end{equation}

\noindent
where $\mathbf{n} \in \mathbb{R}^{h}$ is a learnable unit feature vector for $n_i$, $\mathbf{s_i} \in \mathbb{R}^{n_s}$ and $\mathbf{t_i} \in \mathbb{R}^{n_t}$ are the one-hot vector of $s_i$ and $t_i$ respectively, with $n_s$ and $n_t$ as the number of unique section ids and node (entity) types in the dataset. $\mathit{SciBERT}(z_i) \in \mathbb{R}^{h_e}$ is the hidden representation at the first token ([CLS]) from the final layer of SciBERT.  $\mathbf{W}_s \in \mathbb{R}^{h \times n_s}$ , $\mathbf{W_t} \in \mathbb{R}^{h \times n_t}$, $\mathbf{W}_e \in \mathbb{R}^{h \times h_e}$ are model parameters.  

Following embedding, we contextualize each node representation with 6 GAT layers and pass each node through a final binary classification layer to predict salience.

To supervise the training of this model, we apply the relaxed alignment method to align full graph entities and target graph entities. 
All full graph entities that can be aligned are treated as positive example, and all others as negative. We train with negative sampling by setting the negative sample ratio to be 3. Finally we apply negative loss likelihood function on all positive (labeled as salient) nodes with negative sampling ratio as 3 for training.

\subsection{Complete Quantitative Results}
\label{sec:complete-results}
We provide full qualitative results in this section. We include all evaluation metrics discussed in the paper
for all human test, auto test and valid sets from Table~\ref{full-results-untyped-human} to Table~\ref{full-results-typed-valid}.

\begin{table*}

\begin{minipage}{1.0\textwidth}
\centering

\medskip

\begin{adjustbox}{width=1.0\textwidth}
\begin{tabular}{c c c c c c c | c c c c c c c }

\toprule
\multirow{3}{*}{} & \multicolumn{6}{c|}{\bf Strict Match} & \multicolumn{7}{c}{\bf Relaxed Match}\\
 & Ent P & Ent R & Ent F1 & Rel P & Rel R & Rel F1 & Ent P & Ent R & Ent F1 & Rel P & Rel R & Rel F1 & E Dup \\
\midrule
PR & 15.1 & 27.4 & 18.9 & 4.6 & 6.5 & 5.0 & 22.8 & 38.7 & 27.8 & 6.6 & 9.3 & 7.2 & 1.30\\
TKF &16.6 & 29.7 & 20.7 & 6.1 & 6.4 & 5.7 & 24.3 & 40.7 & 29.4 & 8.9 & 9.4 & 8.4 & 1.41\\
TTG &22.3 & 24.1 & 22.0 & {\bf 9.8} & 6.8 & {\bf 7.2} & 33.9 & 35.3 & 32.7 & {\bf 13.7} & 9.3 & 9.9 & {\bf 1.17}\\
G2G &{\bf 24.1} & {\bf 30.6} & {\bf 24.9} & 7.4 & {\bf 8.3} & 6.9 & {\bf 36.9} & {\bf 42.6} & {\bf 36.9} & 11.3 & {\bf 11.8} & {\bf 10.1} & 1.42\\
\midrule
GE &72.1 & 58.0 & 63.7 & 34.7 & 16.9 & 21.1 & 100.0 & 79.4 & 87.6 & 44.5 & 20.3 & 25.8 & 1.0\\
\bottomrule
\end{tabular}
\end{adjustbox}

\end{minipage}\hfill

\caption{\label{full-results-untyped-human} Full results for \textbf{untyped} entity / relation evaluation (P, R for Precision and Recall) and entity duplication rate (E Dup) on \textbf{human test} set. All scores are in \%, except entity duplication rate. }
\vspace{-3mm}
\end{table*}

\begin{table*}

\begin{minipage}{1.0\textwidth}
\centering

\medskip

\begin{adjustbox}{width=1.0\textwidth}
\begin{tabular}{c c c c c c c | c c c c c c c }

\toprule
\multirow{3}{*}{} & \multicolumn{6}{c|}{\bf Strict Match} & \multicolumn{7}{c}{\bf Relaxed Match}\\
 & Ent P & Ent R & Ent F1 & Rel P & Rel R & Rel F1 & Ent P & Ent R & Ent F1 & Rel P & Rel R & Rel F1 & E Dup \\
\midrule
PR &12.3 & 22.2 & 15.4 & 3.6 & 5.6 & 4.1 & 17.2 & 29.1 & 21.0 & 4.7 & 7.3 & 5.4 & 1.30 \\
TKF &13.2 & 23.6 & 16.5 & 4.4 & 5.1 & 4.2 & 18.0 & 30.0 & 21.8 & 6.1 & 7.1 & 5.9 & 1.41 \\
TTG &18.4 & 19.7 & 18.0 & {\bf 8.2} & 5.7 & {\bf 6.0} & 26.1 & 27.0 & 25.1 & {\bf 11.5} & 7.9 & {\bf 8.3} & {\bf 1.17}\\
G2G &{\bf 19.5} & {\bf 24.8} & {\bf 20.1} & 5.7 & {\bf 6.8} & 5.3 & {\bf 28.0} & {\bf 32.1} & {\bf 27.8} & 8.4 & {\bf 9.3} & 7.6 & 1.42\\
\midrule
GE &61.1 & 49.3 & 54.1 & 27.9 & 14.8 & 17.9 & 79.9 & 64.0 & 70.3 & 34.8 & 17.3 & 21.3 & 1.0\\
\bottomrule
\end{tabular}
\end{adjustbox}

\end{minipage}\hfill

\caption{\label{full-results-typed-human} Full results for \textbf{typed} entity / relation evaluation (P, R for Precision and Recall) and entity duplication rate (E Dup) on \textbf{human test} set. All scores are in \%, except entity duplication rate.}
\end{table*}

\begin{table*}

\begin{minipage}{1.0\textwidth}
\centering

\medskip

\begin{adjustbox}{width=1.0\textwidth}
\begin{tabular}{c c c c c c c | c c c c c c c }

\toprule
\multirow{3}{*}{} & \multicolumn{6}{c|}{\bf Strict Match} & \multicolumn{7}{c}{\bf Relaxed Match}\\
 & Ent P & Ent R & Ent F1 & Rel P & Rel R & Rel F1 & Ent P & Ent R & Ent F1 & Rel P & Rel R & Rel F1 & E Dup \\
\midrule
PR &15.5 & 23.7 & 18.0 & 4.4 & 5.5 & 4.5 & 23.6 & 34.3 & 26.8 & 6.6 & 8.1 & 6.7 & 1.30 \\
TKF &16.4 & 25.3 & 19.1 & 5.2 & 5.6 & 4.9 & 24.6 & 35.4 & 27.9 & 7.8 & 8.1 & 7.3 & 1.34 \\
TTG &{\bf 18.8} & 19.2 & 17.9 & {\bf 7.3} & 4.3 & 5.0 & {\bf 30.2} & 29.7 & 28.3 & {\bf 11.2} & 6.6 & 7.5 & {\bf 1.18}\\
G2G &18.2 & {\bf 30.1} & {\bf 20.8} & 5.5 & {\bf 8.0} & {\bf 5.7} & 29.7 & {\bf 43.8} & {\bf 32.7} & 9.1 & {\bf 12.6} & {\bf 9.2} & 1.55\\
\midrule
GE &69.4 & 56.9 & 62.0 & 34.1 & 15.3 & 19.3 & 100.0 & 81.2 & 88.8 & 44.5 & 18.9 & 24.3 & 1.0\\
\bottomrule
\end{tabular}
\end{adjustbox}

\end{minipage}\hfill

\caption{\label{full-results-untyped-auto} Full results for \textbf{untyped} entity / relation evaluation (P, R for Precision and Recall) and entity duplication rate (E Dup) on \textbf{auto test} set. All scores are in \%, except entity duplication rate. }
\end{table*}

\begin{table*}

\begin{minipage}{1.0\textwidth}
\centering

\medskip

\begin{adjustbox}{width=1.0\textwidth}
\begin{tabular}{c c c c c c c | c c c c c c c }

\toprule
\multirow{3}{*}{} & \multicolumn{6}{c|}{\bf Strict Match} & \multicolumn{7}{c}{\bf Relaxed Match}\\
 & Ent P & Ent R & Ent F1 & Rel P & Rel R & Rel F1 & Ent P & Ent R & Ent F1 & Rel P & Rel R & Rel F1 & E Dup \\
\midrule
PR &12.1 & 18.5 & 14.1 & 3.5 & 4.8 & 3.7 & 17.2 & 24.8 & 19.5 & 4.8 & 6.5 & 5.0 & 1.30  \\
TKF &12.8 & 19.6 & 14.9 & 4.0 & 4.7 & 3.9 & 17.8 & 25.5 & 20.1 & 5.5 & 6.4 & 5.3 & 1.34\\
TTG &{\bf 15.1} & 15.5 & 14.5 & {\bf 6.7} & 3.9 & 4.5 & {\bf 22.7} & 22.3 & 21.2 & {\bf 9.5} & 5.6 & 6.4 & {\bf 1.18}\\
G2G &14.3 & {\bf 23.8} & {\bf 16.4} & 4.4 & {\bf 6.9} & {\bf 4.6} & 21.6 & {\bf 31.8} & {\bf 23.7} & 6.5 & {\bf 10.2} & {\bf 6.9} & 1.55\\
\midrule
GE &57.1 & 47.0 & 51.2 & 27.4 & 13.5 & 16.4 & 77.6 & 63.4 & 69.2 & 34.8 & 16.2 & 20.0 & 1.0\\
\bottomrule
\end{tabular}
\end{adjustbox}

\end{minipage}\hfill

\caption{\label{full-results-typed-auto} Full results for \textbf{typed} entity / relation evaluation (P, R for Precision and Recall) and entity duplication rate (E Dup) on \textbf{auto test} set. All scores are in \%, except entity duplication rate.}
\end{table*}

\begin{table*}

\begin{minipage}{1.0\textwidth}
\centering

\medskip

\begin{adjustbox}{width=1.0\textwidth}
\begin{tabular}{c c c c c c c | c c c c c c c }

\toprule
\multirow{3}{*}{} & \multicolumn{6}{c|}{\bf Strict Match} & \multicolumn{7}{c}{\bf Relaxed Match}\\
 & Ent P & Ent R & Ent F1 & Rel P & Rel R & Rel F1 & Ent P & Ent R & Ent F1 & Rel P & Rel R & Rel F1 & E Dup \\
\midrule
PR &15.7 & 24.3 & 18.4 & 4.5 & 5.8 & 4.6 & 23.7 & 34.6 & 27.1 & 6.7 & 8.3 & 6.8 & 1.27\\
TKF &16.4 & 25.4 & 19.2 & 4.7 & 5.7 & 4.7 & 24.9 & 35.8 & 28.2 & 7.5 & 8.4 & 7.3 & 1.34\\
TTG &{\bf 18.5} & 18.9 & 17.7 & {\bf 7.4} & 4.2 & 4.9 & {\bf 29.9} & 29.5 & 28.2 & {\bf 11.1} & 6.6 & 7.4 & {\bf 1.18}\\
G2G &17.9 & {\bf 30.3} & {\bf 20.9} & 6.0 & {\bf 8.6} & {\bf 6.2} & 29.8 & {\bf 44.5} & {\bf 33.1} & 9.8 & {\bf 13.5} & {\bf 9.9} & 1.52\\
\midrule
GE &68.7 & 56.7 & 61.6 & 35.3 & 16.2 & 20.5 & 99.8 & 81.7 & 89.1 & 45.4 & 19.8 & 25.4 & 1.0\\
\bottomrule
\end{tabular}
\end{adjustbox}

\end{minipage}\hfill

\caption{\label{full-results-untyped-valid} Full results for \textbf{untyped} entity / relation evaluation (P, R for Precision and Recall) and entity duplication rate (E Dup) on \textbf{valid} set. All scores are in \%, except entity duplication rate. }
\end{table*}

\begin{table*}

\begin{minipage}{1.0\textwidth}
\centering

\medskip

\begin{adjustbox}{width=1.0\textwidth}
\begin{tabular}{c c c c c c c | c c c c c c c }

\toprule
\multirow{3}{*}{} & \multicolumn{6}{c|}{\bf Strict Match} & \multicolumn{7}{c}{\bf Relaxed Match}\\
 & Ent P & Ent R & Ent F1 & Rel P & Rel R & Rel F1 & Ent P & Ent R & Ent F1 & Rel P & Rel R & Rel F1 & E Dup \\
\midrule
PR &12.6 & 19.4 & 14.7 & 3.7 & 5.1 & 3.9 & 17.7 & 25.8 & 20.2 & 5.0 & 6.9 & 5.3 & 1.27 \\
TKF &13.0 & 20.0 & 15.2 & 3.6 & 4.9 & 3.8 & 18.4 & 26.4 & 20.9 & 5.3 & 6.9 & 5.4 & 1.34\\
TTG &{\bf 15.1} & 15.6 & 14.6 & {\bf 6.9} & 3.9 & 4.5 & {\bf 22.6} & 22.4 & 21.3 & {\bf 9.6} & 5.6 & 6.3 & {\bf 1.18}\\
G2G &14.1 & {\bf 24.0} & {\bf 16.5} & 4.8 & {\bf 7.7} & {\bf 5.2} & 21.8 & {\bf 32.6} & {\bf 24.3} & 7.1 & {\bf 11.1} & {\bf 7.6} & 1.52\\
\midrule
GE &56.6 & 46.8 & 50.8 & 28.6 & 14.5 & 17.6 & 77.1 & 63.4 & 69.0 & 36.0 & 17.3 & 21.2 & 1.0\\
\bottomrule
\end{tabular}
\end{adjustbox}

\end{minipage}\hfill

\caption{\label{full-results-typed-valid} Full results for \textbf{typed} entity / relation evaluation (P, R for Precision and Recall) and entity duplication rate (E Dup) on \textbf{valid} set. All scores are in \%, except entity duplication rate.}
\end{table*}

\subsection{Implementation Details}
\label{sec:implementation-details}

\paragraph{DyGIE++} We do not re-train DyGIE++, instead we use the pretrained model on SciERC for all processing and modeling steps in this work.

\paragraph{TTG} We apply the same hyperparameters from BertSumExt for finetuning on all (full paper, abstract) pairs in our train set. In order to make the model better fit into the scientific domain, we replace the original pretrained BERT base model with the SciBERT base model. And as papers in our datasets are generally long, we increase the maximum sequence length of the original model from 512 to 1024. 

\paragraph{G2G} We manually tune the hyperparameters of GAT based on dev set entity F1 performance curve versus training steps. We fix most of the model parameters (e.g. vector dimension $h$ = 16; number of layers = 6; batch size = 10). The only parameters being tuned are learning rate, dropout rate and negative sampling ratio. But we only manually change each parameter value if we observe performance instability on the dev set for the first 1000 training steps. The average number of tuning trials for each parameter is fewer than 5 times. Finally we set negative sampling ratio = 3, dropout rate = 0.2 and learning rate = 5e-5 for all experiments with our G2G model. We use Adam optimizer. We do not finetune SciBERT (the base model) used in GAT. 

We run each experiment on a single TITAN RTX. We select the model checkpoint based on its typed relation F1 score performance on valid set.